\documentclass[sigconf]{acmart}

\AtBeginDocument{%
  \providecommand\BibTeX{{%
    \normalfont B\kern-0.5em{\scshape i\kern-0.25em b}\kern-0.8em\TeX}}}

\copyrightyear{2021}
\acmYear{2021}
\setcopyright{acmlicensed}\acmConference[ICAIF'21]{2nd ACM International Conference on AI in Finance}{November 3--5, 2021}{Virtual Event, USA}
\acmBooktitle{2nd ACM International Conference on AI in Finance (ICAIF'21), November 3--5, 2021, Virtual Event, USA}
\acmPrice{15.00}
\acmDOI{10.1145/3490354.3494448}
\acmISBN{978-1-4503-9148-1/21/11}

\usepackage{algorithm}
\usepackage{algorithmic}
\begin{document}

\title{Deep Q-Learning Market Makers in a Multi-Agent Simulated Stock Market}


\author{Óscar Fernández Vicente}
\affiliation{%
  \institution{Universidad Carlos III Madrid}
  \city{Madrid}
  \country{Spain}}
\email{oscar.f.vicente@alumnos.uc3m.es}

\author{Fernando Fernández Rebollo}
\affiliation{%
  \institution{Universidad Carlos III Madrid}
  \city{Madrid}
  \country{Spain}}
\email{ffernand@inf.uc3m.es}

\author{Francisco Javier García Polo}
\affiliation{%
  \institution{Universidad Carlos III Madrid}
  \city{Madrid}
  \country{Spain}}
\email{fjgpolo@inf.uc3m.es}

\begin{abstract}

Market makers play a key role in financial markets by providing liquidity. They usually fill order books with buy and sell limit orders in order to provide traders alternative price levels to operate. This paper focuses precisely on the study of these markets makers strategies from an agent-based perspective. In particular, we propose the application of Reinforcement Learning (RL) for the creation of intelligent market markers in simulated stock markets. This research analyzes how RL market maker agents behaves in non-competitive (only one RL market maker learning at the same time) and competitive scenarios (multiple RL market markers learning at the same time), and how they adapt their strategies in a Sim2Real scope with interesting results. Furthermore, it covers the application of policy transfer between different experiments, describing the impact of competing environments on RL agents performance. RL and deep RL techniques are proven as profitable market maker approaches, leading to a better understanding of their behavior in stock markets.  

\end{abstract}

\keywords{deep learning, reinforcement learning, multi-agent, stock markets, market makers}


\maketitle

\section{Introduction}

\begin{figure}[ht]
        \centering
        \includegraphics[width=0.5\textwidth]{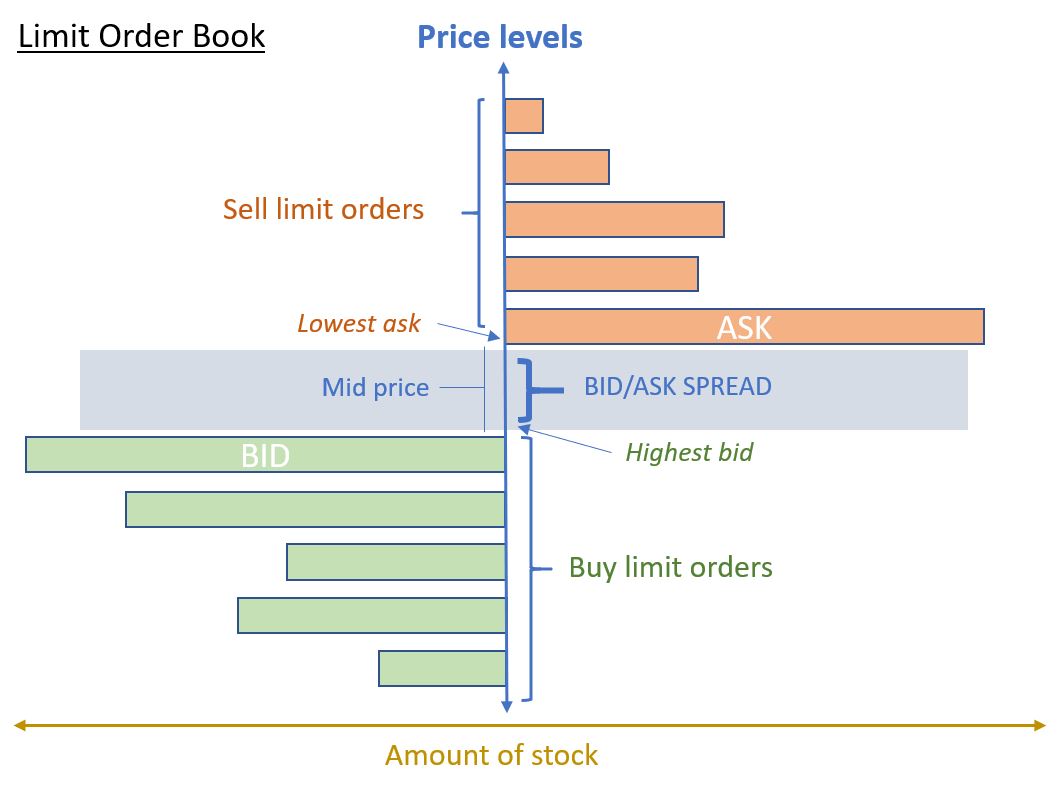}
        \caption{Limit order book structure and concepts}
        \label{fig:lob}
        \end{figure}

Stock markets are environments where many kind of participants, with different goals, interact by buying and selling stocks. There is an auction where agents stream buy and sell prices to a public centralised limit order book (OB) in order to be matched by other agents~\cite{Lu2018}. \autoref{fig:lob} illustrates the main concepts of every limit order book. When two agents deal to buy and sell at the same price, transaction is executed. This is how stock markets work in a simplified way. In stock markets there is also one kind of agents called market makers, which main objective is to provide liquidity while reducing risk and earning some profits. They usually quote multiple buy and sell orders at different price levels, in order to give traders more options to execute match their trades~\cite{Chakraborty2011}. This is quite important especially in markets where there is certain lack of liquidity. Market makers have to be very careful managing their inventory while carrying out this task, since having large amount of stocks can cause big losses due to price volatility. This trade off between inventory and quoted bid/ask prices (spread) must be optimized in order to be as much profitable as possible while reducing risks.

One of the alternatives to study these financial markets is the use of simulators that allow us to recreate the market conditions. In these simulated markets, machine learning~\cite{mitchell}, and especially reinforcement learning (RL) techniques~\cite{sutton2011} are increasingly reliant for agent training and testing.
RL is a promising technique where agents try to gain the maximum long term reward from an environment while performing a specific task. They learn, by trial an error, how to get the best way to reach some goal. RL has been applied in multiple domains, such as video-games~\cite{Shao2019}, robotics~\cite{Kober2013} and also stock markets~\cite{Fischer2018}. 

Bearing all these things in mind, the goal of this paper is to study the behavior of the market markers from an agent-based perspective, where we propose the application of RL for the creation of intelligent market makers. In particular, this paper aims to define an environment where RL market makers compete with other predefined market makers agents, and also among them, to achieve the most profitable strategy. According to this, it is intended to analyze the behavior of market makers from a competitive and non-competitive point of view that allows a better understanding of the trading strategies that they achieve when they trade alone and when they do against other RL market makers. 

Market makers strategies have been broadly studied not only from a classical approach~\cite{Avellaneda2008}, but even using cutting-edge techniques, including RL. From this RL perspective, Chang and Shelton~\cite{Chan2001} launch different RL strategies (SARSA, actor-critic); Yagna Patel~\cite{Patel2018} combines two types of agents (micro/macro) with different inputs to achieve a profitable strategy; Spooner~\cite{Spooner2018} focuses on building a realistic simulation of an order book with Q-learning agents. Related to market maker in multi-agent environments, there are also some interesting research. For instance, Wang, Zhou and Zeng~\cite{Wang2011} run multiple experiments on a simulated markets with informed an uninformed traders, focusing on inventory optimization. Recently, in 2019, Ganesh, Vadori, Xu, Zheng, Reddy and Veloso ~\cite{Ganesh2019} focus on a competitive multi-agent scenario, using ABIDES as simulation environment, where RL agents try to beat other ramdom market maker agents by finding profitable policies. Although previous papers focus on using RL agents from a profitability and risk management perspective, this paper aims to focus also on the interaction among them, analysing the impact of this interaction in terms of profitability and understanding how different strategies are taken by competing market maker agents in single-agent and multi-agent environments. Furthermore, pre-trained RL market makers are run in the same experiment using direct transfer learning to evaluate their performance and policies. 

The experiments are conducted using ABIDES (Agent-Based Interactive Discrete Event Simulation environment)~\cite{Byrd2019} which simulates a Financial Market. It is inside this environment where experimental market makers interact with investor agents that always pick up the cheapest market marker in terms of spread. From a financial perspective, the experiments show that, when RL market makers interact against simple agents, they are much more profitable than when other RL market makers participate simultaneously. It is remarkable how RL market makers adapt their strategies dynamically to adapt to more challenging scenarios, where multiple RL agents take place. These insights can be useful in order to define further profitable strategies based on machine learning, not only from market maker perspective but also from stock trading point of view.

The organization of this paper is as follows: Section~\ref{sec:background} describes RL from academic point of view. 
Section~\ref{sec::mapping} introduces the task of applying RL on financial markets, defining states, actions and rewards. It also represents the architecture of RL agents.
Section~\ref{sec:singlemulti} show the theory behind the experiments. 
Section~\ref{experiments} describes all the experiment launched, with the results obtained.
At last section~\ref{conclusions}, main conclusions obtained from the research are discussed.

\section{Background on Reinforcement Learning}
\label{sec:background}

Reinforcement learning is a machine learning technique where agents usually find the best way to do a task by optimizing the rewards earned after every action. These tasks are usually described as Markov Decision Processes (MDP). MDPs are represented by the following tuples $\mathcal M=\langle S, A, T, R \rangle$, where $S$ corresponds to the state space representation, $A$ to the action space, $T$ to the transition function between states, and finally $R$ corresponds to the immediate rewards earned after every action taken. The main goal of the RL agents consists on selecting the best policy $\pi^*$ that maximizes the following return:\newline
\begin{equation}
    G_t = \sum_{k=0}^{\infty}\gamma^k R_{t+k+1}
\label{eq:sumrewards}
\end{equation}
where $\gamma$ is a parameter called the discount rate, 0 $\le$ $\gamma$ $\le$ 1. This discount rate represents the value assigned to future rewards. The closer to 0, the more importance given to immediate rewards.
The function that defines the sum of rewards obtained being in one state with some policy is called the value function, $V^{\pi}(s)$. Also we can determine the value function of being in some state and taking some action $Q^{\pi}(s,a)$ according to policy $\pi$. There are some algorithms that computes the action-value functions. Probably, Q-learning algorithm~\cite{watkins1989} is one of the most widely used. It basically relies on the use of a table that holds the estimated discounted reward of every state and action pair. This technique can be used in small domains, with discrete state and action spaces. When the state or action spaces is very large, or we have continuous domains, some other techniques are used alternatively. This is the case of Deep Q-Network (DQN) \cite{Mnih2015}, a deep learning technique that uses a neural network instead of table in order to approximate the value function.

\section{Mapping a Market Market task onto a Reinforcement Learning task}
\label{sec::mapping}

In this section, we describe the mapping of the trading problem of a market marker onto a RL task. For this purpose, first the state space $S$, the action space $A$, and the reward function $R$ are explained in order to have a better understanding of the system.  We assume the transition function $T$ is unknown for the learning agent. After that the architecture of the Deep Q-Network agent and its neural network and the algorithm below the learning process are also represented.

\subsection{State and Action Space}

\textbf{States}: The observation state consists of 10 features: (i) number of buy operations executed in the previous time step, (ii)  number of stocks bought in the previous time step, (iii) number of sell operations executed in the previous time step, (iv) number of stocks sold in the previous time step, (v) inventory size (positive or negative) in the previous step, (vi) current inventory size, (vii) stock mid price variation between time steps ($P_{var} = current mid price - last mid price$), (viii) current reference spread, (ix) spread in previous time step, and (x) total amount of stock traded in previous time step, typically known as $volume$. This space representation has been chosen as most relevant because it has every important element that market provides with. It describes not only the current spread and inventory, but also what happened in previous time step. It also shows the evolution of mid price (important in terms of inventory value). Other approaches could have been taken instead, such as the use of rolling windows averages or relative differences of some input values. The search of more accurate state space variables and its optimization could be addressed as another line of work.

\textbf{Actions}: There are three decision variables that market makers can interact with in every time step: (i) buy spread, (ii) sell spread and (iii) amount of inventory to hedge. Therefore, our action space is composed of these three variables. In this case, we have opted for "discretizing" them. We have opted for this solution instead of continuous action space in order to increase convergence rate, sacrificing an exact value which is not so important. Buy and sell actions will consist of picking one epsilon in the following list of evenly distributed values: [-1, -0.8, -0.6, ... 0.6, 0.8, 1].

Therefore, streamed buy/sell spreads will consist of the following formula: $Spread_{mm}$ = $Spread_{ref}$ * ($1$+$epsilon$), where $Spread_{ref}$ is the current price spread streamed by the rest of agents (\autoref{subsec:experimentalsetting}) and $Spread_{mm}$ is the spread that market maker is going to stream. On the other hand, hedging action will consist on picking another epsilon within list [0 , 0.25, 0.5, 0.75, 1]. In this way, market makers will reduce their inventory according to the selected epsilon, from 0\% to 100\% on every time step, according to $inv$ = $inv$ * $epsilon$, where $inv$ is the current inventory held by the market marker. Buy and sell epsilons can be same or different, incurring in skew. After applying this discretization process, the action space reduces to 605 actions, corresponding to 11 buy's epsilon $\times$ 11 sell's epsilons $\times$ 5 hedge's epsilons.

\textbf{Rewards}: Total reward obtained in every time step will consist of the following Equation:

\begin{equation}
    r = b/s + invPnL- hedCost
\end{equation}

\noindent where:

\begin{itemize}
\item (i) $b/s$: It corresponds to the amount of stock traded, buys or sells. For every transaction made the market makers earns amount $b/s = num_{b/s} * Spread_{mm}$, where $num_{b/s}$ is amount of stocks traded, and $spread_{mm}$ is the spread quoted by market maker.
\item (ii) $invPnL$: Inventory held by the market marker will earn or lose value according to mid-price variation in every time step. If price rises their inventory value will rise as well, and vice versa. It is computed as: $invPnL = inv * P_{var}$, where $inv$ is the amount of stock held by market maker, and $P_{var}$ is the difference between current stock price and last time step price.

\item (iii) $hedCost$: Every time step a market marker hedges (reduce) some of its inventory, it pays a cost for this action. This cost would be the amount of stock executed by the $Spread_{ref}$, as it is traded at last market price. It is computed as $hedCost =  InvHedged_{mm} * Spread_{ref}$, where $InvHedged_{mm}$ is the amount of stocks hedged and $Spread_{ref}$ is the current market spread.
\end{itemize}

The proposed state and action space, and the reward function will be used for the RL market market agent to learn a trading strategy. 

\subsection{Deep Q-Learning Market Maker Agents (DQL-MM)}
\label{sec:dqlmm}

\begin{figure}[ht]
        \centering
        \includegraphics[width=0.5\textwidth]{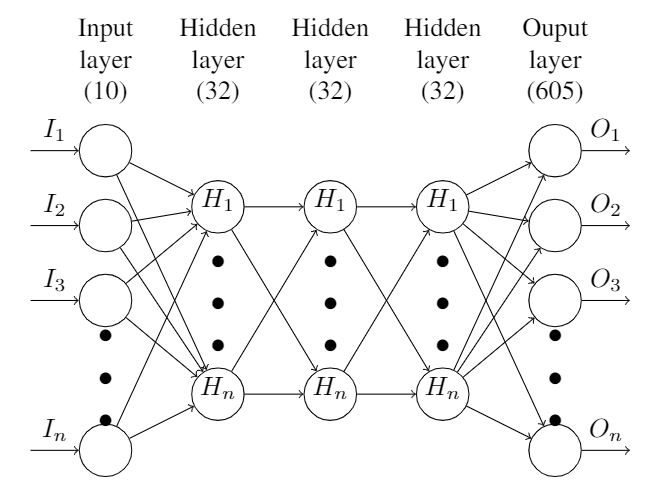}
        \caption{Deep Neural Network architecture used for the DQL-MM agent.}
        \label{pic:nn}
        \end{figure}

Due to the continuous and high-dimensional nature of the proposed RL task, in this paper we have opted for using Deep Learning strategies to learn the behavior policy of the RL market makers. Every RL market maker has been based on a DQN (Deep Q-Network) architecture \cite{Mnih2015}. These DQN market maker agents (DQL-MM) are conformed by a fully-connected neural network (NN) which predicts the expected reward taking an action in current time step. This NN adjusts its weights every 200 time steps. The NN architecture defined for the experiments is as follows: It has an input  layer with 10 nodes, three hidden layers with 32 nodes each, and one output layer with 605 nodes, i.e., one output node for every possible action. Figure~\ref{pic:nn} shows a graphical representation of the proposed Deep Q-Network architecture.

All three hidden layers have a ReLu activation function, while output layers has a linear activation one. Adam has been chosen as optimizer and MAE as the loss metric. Input data has been previously standardized using StandardScaler from Scikit-learn library \cite{Scikit-Learn}.  Each output represents a combination of one buy's epsilon, one sell's epsilon and one hedge's epsilon. Keras for Tensorflow has been used for the NN model.

Algorithm \autoref{alg:example} shows the learning process of DQL-MM agents, and it is based on experience replay \cite{Mnih2015}. For every simulations considered, DQL-MM import NN from last simulation (line 5 in Algorithm \autoref{alg:example}). Once a simulation starts, along all time steps DQL-MM queries current reference market spread (line 8) $Spread_{ref}$ that comes from interaction of the rest of agents (Section 6.1). After that, it is time to pick an action. According to the $\epsilon$-greedy strategy and the value of the parameter $\epsilon$, selected action $a_{t}$ will be the best possible (greedy) (line 11) or a random one (line 13). Best possible action will be returned by the NN that computes the value function of every possible action for that state. DQL-MM calculates rewards earned the previous time step according to taken actions, and the tuple $\langle s_{t},a_{t}, s_{t+1},r_{t}\rangle$ is stored to a buffer $\mathcal D$ [17]. Every 200 time steps, NN is retrained with stored tuples in $\mathcal D$ (line 20).

\begin{algorithm}[tb]
   \caption{DQL-MM flow}
   \label{alg:example}
\begin{algorithmic}[1]
    \STATE Initialize memory $\mathcal D \leftarrow \emptyset$, $t \leftarrow 0$
    \FOR{$simulation$}
   \STATE Init simulation
   \IF{Saved agent state}
   \STATE Get last NN weights
   \ENDIF

   \FOR{$every \ time \ step \ in \ simulation$}
   \STATE Get  $Spread_{ref}$ from market data.
   \STATE Initialise state, $s_{t}$
   
   \IF{$rand < \epsilon$}
   \STATE $a_{t}$=best\_action($buy$, $sell$ and $hedge$ $epsilons$)
   \ELSE
   \STATE $a_{t}$=random\_action($buy$, $sell$ and $hedge$ $epsilons$)
   \ENDIF
   \STATE Execute action $a_{t}$, observe $r_t$ and $s_{t+1}$
  \STATE Apply $\epsilon$ decay
   \STATE Store transition $\langle s_{t}, a_{t}, s_{t+1}, r_{t} \rangle$ buffer in $\mathcal D$
   \STATE $t \leftarrow t + 1$
   \IF{$t$  $\%$ $200$ == $0$}
   \STATE Retrain NN with transitions in $\mathcal D$ 
   \STATE $\mathcal D \leftarrow \emptyset$
   \ENDIF
   \STATE $s_{t} \leftarrow s_{t+1}$
   \ENDFOR
   \ENDFOR
\end{algorithmic}
\end{algorithm}
\section{Single-agent and Multi-agent scenarios}
\label{sec:singlemulti}

Two main environments are studied in this paper. First of all, it is interesting to evaluate how a single DQL-MM agent behaves in a stationary\cite{Gronauer2021} competitive trading domain, with two other non-RL MM agents (\autoref{subsec:learningsingle}). How good or bad a DQL-MM agent performs against other non-RL MM agents in terms of profitability and stability, and the importance of the variables involved in the decision making process. Once this case is studied, it is specially relevant to evaluate how DQL-MM agents behaves while competing among other DQL-MM agents, in an environment perceived as non-stationary by every single-agent (\autoref{subsec:learningmultiple}). In this more complex scenario, market makers have to deal with dynamic strategies from other competitors, by adapting their actions in order to earn the maximum reward. In this multi-agent environment, policy transfer is also evaluated in order to check the performance of winner pre-trained policies in new scenarios (\autoref{subsec:transfer}). This can be really useful if we want to transfer winner strategies from simulated environments such as ABIDES, to real markets. In the following experiments we analyse both approaches (single and multi-agent) and give us interesting insights about market maker behaviours, strategies and profitability. 

\section{Experiments}
\label{experiments}

This section evaluates the proposed DQL-MM agents for the learning of market making strategies in different scenarios. In particular, three configurations are launched in this experimental section. First two of them are related to learning best strategies in two different scenarios: (i) a single DQL-MM agent competing againts two pre-defined non-RL market makers (Section~\ref{subsec:learningsingle}), and (ii) three DQL-MM agents competing each other (Section~\ref{subsec:learningmultiple}), also with two more non-RL market maker agents. (iii) The third experiment relies on the transfer of pre-trained market maker policies in a competitive scenario (Section~\ref{subsec:transfer}). \newline The experimental setting is described first in the following Section~\ref{subsec:experimentalsetting}.

\subsection{Experimental Setting}
\label{subsec:experimentalsetting}

Each simulation launched in ABIDES is based on a 2 hour's market session, run with the following agents: 

\begin{itemize}

\item {100 Noise agents}: This agents place orders in a random direction of fixed size.
\item {10 Value agents}. They have access to fundamental time series and operate according to variation of mid-price related to their mid-price forecast. 
\item {10 Momentum agents}. This kind of agents operate when 50 and 20 steps moving averages cross.
\item {1 Adaptive POV Market Maker Agent}. This agent provides certain liquidity to the synthetic market by placing orders at fixed intervals.
\item {1 Exchange agent}. Finally, this agent orchestrate the interaction and integration of all the agents and order book.
\end{itemize}

These agents interact during this two market hours defining a reference spread ($Spread_{ref}$), and certain price volatility according to their trading activity. It is in this environment where experimental market makers activity takes place. There are three kinds of  experimental market makers:

\begin{itemize}
\item \textbf{DQL market maker} (DQL-MM): Deep Q-Learning (DQN) market maker.
\item \textbf{Random market maker} (Random-MM), with randoms spreads and hedges on every time step.
\item \textbf{Persistent market maker} (Persistent-MM), with fixed random spread and hedges over every simulation.
\end{itemize}

These market maker agents interact with 50 extra investors agents. Investor agents place buy or sell orders in an uniform random way. They always matches their orders against the market maker with the narrowest spread (buy or sell respectively), hence the cheapest one. In case of being found same spreads among market makers, investors will take one of them randomly to avoid bias produced by MM order. No transaction fees have been considered in any trade.

Basic RL parameters of DQL agents are the following: starting $\epsilon$: 0.99, $\epsilon$ decay at every time step: 0.99999, minimum $\epsilon$: 0.01. $\gamma$: 0.6.

A total of 250 simulations were run on every experiment. Also, 5 rounds have been executed in parallel to reduce the stochastic uncertainty in results. All the simulations have been run in a Intel i7-9700 Linux (WSL) machine with 32GB RAM.

\subsection{Single-agent market maker}
\label{subsec:learningsingle}

\begin{figure}[ht]
        \centering
        \includegraphics[width=0.5\textwidth]{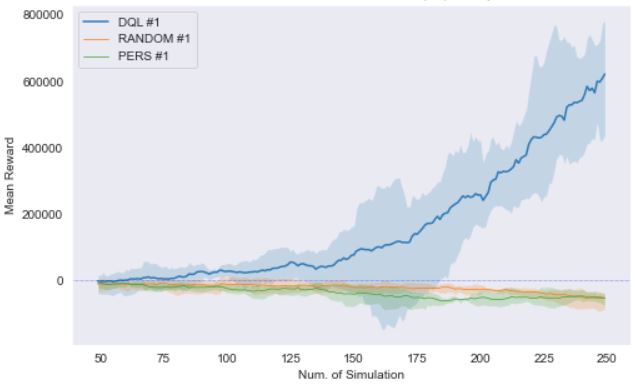}
        \caption{Single-agent experiment - (50 sims. rolling window).}
        \label{fig:exp1}
        \end{figure}

In this first experiment, a single DQL-MM agent competes against two other non-RL market makers: one Random-MM and one Persistent-MM. \autoref{fig:exp1} shows the evolution in performance of the three market makers agents along the experiment. It is noticed how DQL-MM agents gets positive rewards from early stages of experiment, comparing to other two agents. However, it is at simulation 175 when performance increases significantly. 

\autoref{table-1} shows mean rewards of 250 simulations (all) obtained by every experimental market maker on every simulation. On one hand both Random-MM and Persistent-MM agents end the experiment with negative rewards in average. On the other hand, DQL-MM is able to learn a good policy that clearly outperforms former 2 agents, in terms of earnings. 

\begin{table}[t]
\caption{Single-agent experiment average results (250 exp) (USDx10$^3$)}
\label{table-1}
\vskip 0.15in
\begin{center}
\begin{small}
\begin{sc}
\begin{tabular}{lcccr}
\toprule
Data set & MEAN & TOP & BOTTOM & STD \\
\midrule
DQL-MM    & 159   & 782 & -148& 180\\
RANDOM-MM    & -23& 13  & -94 & 11  \\
PERS-MM    & -37& 0 & -91& 15 \\
\bottomrule
\end{tabular}
\end{sc}
\end{small}
\end{center}
\vskip -0.1in
\end{table}

\begin{figure}[ht]
        \centering
        \includegraphics[width=0.5\textwidth]{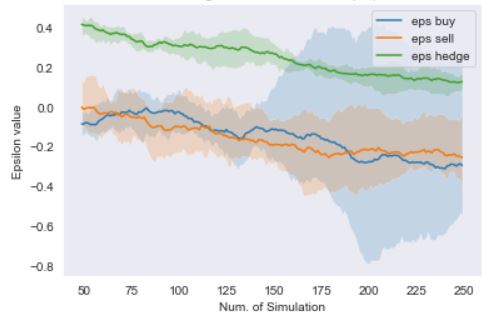}
        \caption{Epsilons evolution along single-agent experiment - (50 sims. rolling window).}
        \label{fig:eps1}
        \end{figure}

\autoref{fig:eps1} shows how buy, sell and hedge epsilons evolve along all the experiment. As mentioned before, these three variables are the possible actions that MM have to deal with. All three epsilons are 50 simulations moving average. This kind of figures illustrates how DQL-MM strategy adapts over time. Buy and sell epsilons tend to converge close to $-0.2$ value, as shown in \autoref{fig:eps1}. However buy epsilon increases its volatility from simulation number 150, probably finding an optimum strategy. Negative buy/sell epsilon values, as current ones, means cheaper prices than $Spread_{ref}$. The smaller the buy/sell epsilon, the more probable to be matched by an investor, but getting fewer rewards per trade. Hedge epsilon converges close to $0.2$, what means that DQL remove the 20\% of its inventory every time step, incurring in hedging costs (as it is traded at $Spread_{ref}$).
 
\begin{figure}[ht]
        \centering
        \includegraphics[width=0.5\textwidth]{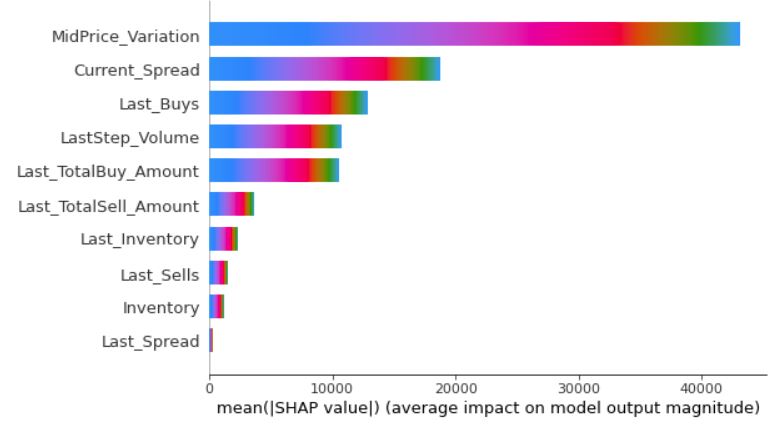}
        \caption{SHAP variable importance summary plot.}
        \label{fig:shap}
        \end{figure}

\textbf{Input variable analysis}:      
Taking a deeper look at DQL-MM single-agent, which tends to perform better than the rest of agents, despite volatility, it is possible to extract more relevant information from its neural network. By using a library such as SHAP \cite{Shap} we can analyse what variables have more importance in order to determine the final policy and strategy. This can be useful not only from a variable optimization perspective, but from the understanding of the related market maker policy and inputs. According to this, \autoref{fig:shap} shows the weights of different space state variables in final action selection. The longer the size of the bar, the more relevant is the variable on model. The color of the bar is related to the final action selected, as each color represents one specific action. As we can see in the graph $P_{var}$ (MidPrice\_Variation) is specially relevant when determining an action. The second most important variable is $Spread_{ref}$ (current\_spread), while variables related to $inv$ (Last\_Inventory and Inventory) and $Spread_{ref}$ from last time step (LastStep\_Spread) have low impact on model.

\subsection{Multi-agent RL market makers}
\label{subsec:learningmultiple}

\begin{figure}[ht]
        \centering
        \includegraphics[width=0.5\textwidth]{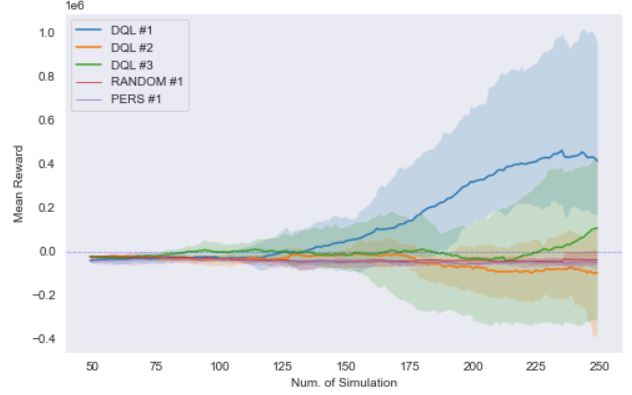}
        \caption{Multi-agent experiment - (50 sims. rolling window).}
        \label{fig:exp2}
        \end{figure}

In this second experiment two more RL agents are included. Hence, a total three similar DQL-MM agents are competing each other in the same environment, also against one Random-MM and one Persistent-MM. \autoref{fig:exp2} illustrates how just the first DQL-MM \#1 is able to achieve good returns, something more remarkable since simulation number 175. DQL-MM \#3 seems to achieve a profitable strategy at last steps, degrading DQL-MM \#1 performance. DQL-MM \#2 however can not discover a competitive strategy along the experiment, ending in red figures.

\begin{figure}[ht]
        \centering
        \includegraphics[width=0.5\textwidth]{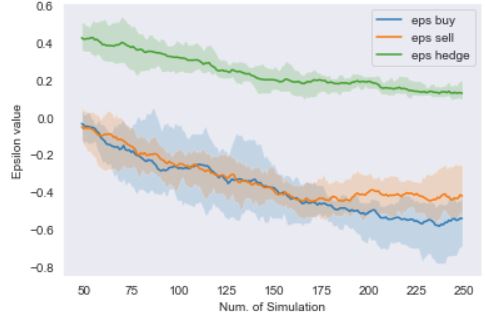}
        \caption{DQL-MM \#1 Epsilons - (50 sims. rolling window).}
        \label{fig:eps2a}
        \end{figure}

\begin{figure}[ht]
        \centering
        \includegraphics[width=0.5\textwidth]{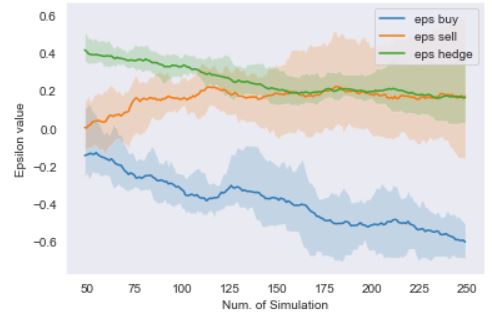}
        \caption{DQL-MM \#2 Epsilons - (50 sims. rolling window).}
        \label{fig:eps2b}
        \end{figure}

\begin{figure}[ht]
        \centering
        \includegraphics[width=0.5\textwidth]{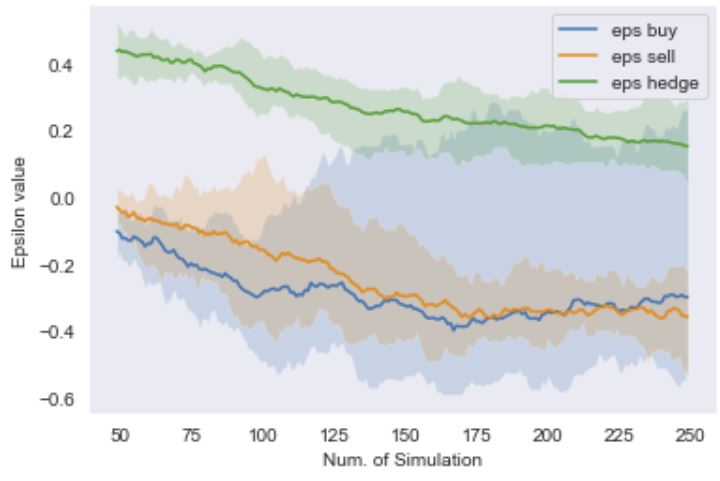}
        \caption{DQL-MM \#3 Epsilons - (50 sims. rolling window).}
        \label{fig:eps2c}
        \end{figure}

As shown in \autoref{table-2}, just the first DQL-MM \#1 agent is able to end the experiment with positive returns in average. Every agent learn different strategies along the experiment, trying to adapt their actions to the most rewarding policy. This strategy evolution is described in figures \autoref{fig:eps2a} (DQL-MM \#1), \autoref{fig:eps2b} (DQL-MM \#2) and \autoref{fig:eps2c} (DQL-MM \#3). Each figure shows how buy, sell and hedge epsilons evolve along the experiment for RL agents individually. 

\begin{table}[t]
\caption{Multi-agent experiment average results (250 exp) (USDx10$^3$)}
\label{table-2}
\vskip 0.15in
\begin{center}
\begin{small}
\begin{sc}
\begin{tabular}{lcccr}
\toprule
Data set & MEAN & TOP & BOTTOM & STD \\
\midrule
DQL-MM \#1   & 129  & 1.018  & -94 & 179 \\
DQL-MM \#2   & -45 & 111  & -393  & 27 \\
DQL-MM \#3   & -5 & 434  & -343  &  24 \\
RANDOM-MM    & -38  & 9& -87  &  7 \\
PERSISTENT-MM    & -45  & 2  & -85 & 6  \\
\bottomrule
\end{tabular}
\end{sc}
\end{small}
\end{center}
\vskip -0.1in
\end{table}

As noticed in mentioned figures, buy and sell epsilons change among three DQL-MM agents, as they continuously adapt their strategies to the changing environment. It should be noted that, as it is a zero-sum game, every strategy taken by an agent impacts directly on the rest agents. The better one agent performs, the worse the rest will do. According to this assumption they evolve their strategies in three different ways. DQL-MM \#1 (\autoref{fig:eps2a}) has a similar epsilon's strategy as DQL-MM in experiment 1 (single DQL-MM agent), with less variance in the whole experiment. DQL-MM \#3 strategy (\autoref{fig:eps2c}) is very similar to DQL-MM \#1 one, but clearly increasing buy epsilon volatility testing new policies. Both average buy and sell epsilons converge close to $-0.4$. DQL-MM \#2 strategy, however, (\autoref{fig:eps2b}) differs from other two agents by moving its sell epsilons toward positive figures.

Hedging epsilons are very similar in all three DQL-MM, converging smoothly to $0.2$. This hedging epsilon seems to be the most optimal in this environment too. \newline We can conclude that when multiple RL agents take place in the same experiment, they suffer much more due to the competitive environment, although they try to adapt their policies looking for being profitable.

\subsection{Direct transfer learning in multi-agent RL market makers}
\label{subsec:transfer}

\begin{figure}[ht]
        \centering
        \includegraphics[width=0.5\textwidth]{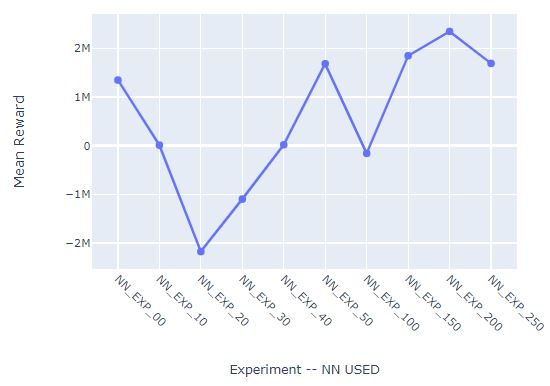}
        \caption{NN Performance evaluation. Mean rewards obtained by every NN tested.}
        \label{fig:nnexp}
        \end{figure}

Once executed the experiments of Sections~\ref{subsec:learningsingle} and~\ref{subsec:learningmultiple}, it is interesting to see the performance of both DQL-MM winner agents competing in the same environment. The main idea behind this is to evaluate if there is some kind of pre-trained policy better than the rest of them, that could be directly transferred to new experiments. This idea has been evaluated using direct transfer learning, by applying neural networks from both "best performance" DQL-MM agents. Regarding to the first experiment of Section ~\ref{subsec:learningsingle}, and in order to use the best learned policy, it has been evaluated the performance of neural networks obtained at 10 different simulations: $[0, 10, 20, 30, 40, 50, 100, 150, 200, 250]$. The results of this performance evaluation has been plotted at figure \autoref{fig:nnexp}. As shown in this figure, the NN with best performance was related to simulation 200, therefore this NN was the transferred one.

\begin{figure}[ht]
        \centering
        \includegraphics[width=0.5\textwidth]{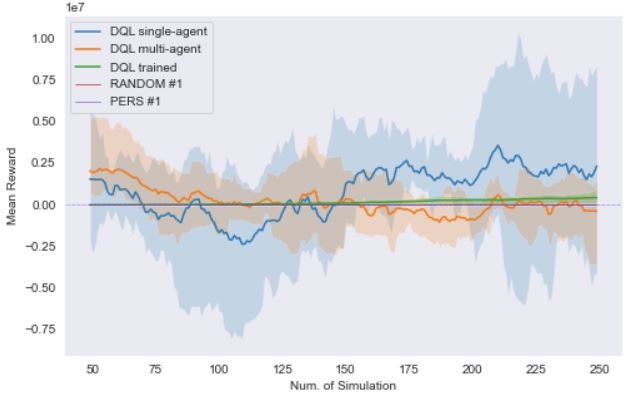}
        \caption{Direct transfer learning experiment results - (50 sims rolling window).}
        \label{fig:tlexp}
        \end{figure}

Regarding to the second experiment of Sections~\ref{subsec:learningmultiple} it was transferred the same NN, also corresponding to simulation 200. Both pre-trained DQL agents were run in a new environment, competing against another learning DQL-MM agent, plus an additional Random-MM and a Persistent-MM agent. \autoref{fig:tlexp}, as prevoius experiments, shows the rewards obtained by every MM along the experiment. 

The results shows some interesting insights. On the one hand pre-trained DQL-MM single-agent from first experiment (blue) behaves better than other DQL-MM, in terms of average earnings. On the other hand, pre-trained DQL multi-agent from second experiment (orange) also has good returns, not as good as first one, but with less volatility. This multi-agent pre-trained policy seems to be less risky than single-agent one. Both pre-trained DQL-MM agents performs very well before simulation 175-200. It is at this point where the learning DQL-MM agent (green) improves its results, as seen in former experiments. This improvement produces a downgrade in rewards earned by multi-agent DQL-MM agent (orange). 

\begin{figure}[ht]
        \centering
        \includegraphics[width=0.5\textwidth]{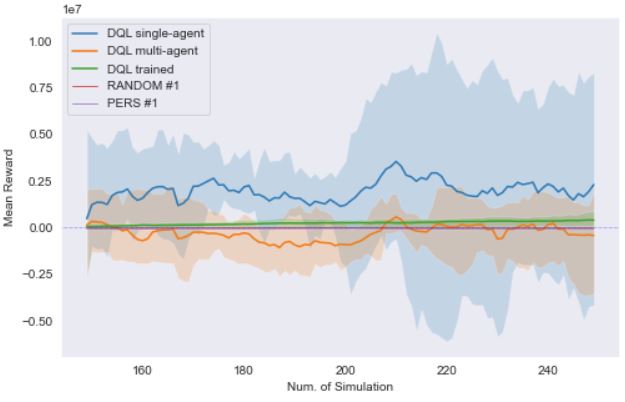}
        \caption{Last simulations transfer learning results - (50 sims. rolling window).}
        \label{fig:nnexp_zoom}
        \end{figure}

Moreover it is noticed that, from simulation 200, pre-trained DQL-MM single-agent increases significantly its variance in results as seen in  \autoref{fig:nnexp_zoom}. This figure focuses on last 100 simulations of \autoref{fig:tlexp}. Random-MM and Persistent-MM agents performance continue being negative, as in previous experiments. 
Lastly, it is important to remark that learning DQL-MM is not as much profitable as it was in experiment one, where it was the only RL market maker agent. Moreover, training DQL-MM (green) volatility remains very low compared to pre-trained DQL-MM agents, what indicates that it is less risky and more consistent to have learning agents than fixed pre-trained ones, no matter how good the pre-trained policy is. \autoref{table-3} shows final rewards obtained by every agent in average.

\begin{table}[t]
\caption{Direct transfer learning experiment average results (250 exp) (USDx10$^3$)}
\label{table-3}
\vskip 0.15in
\begin{center}
\begin{small}
\begin{sc}
\begin{tabular}{lcccr}
\toprule
Data set & MEAN & TOP & BOTTOM & STD \\
\midrule
DQL-MM single   & 780  & 10.447 & -8.187  & 1.442 \\
DQL-MM multi   &165  & 5.268  & -3.622 & 742 \\
DQL-MM trained   & 121  & 877  & -76 & 138 \\
RANDOM-MM    & -37 & 15  & -64   & 5  \\
PERSISTENT-MM    & -50  & -15  & -90  & 5 \\
\bottomrule
\end{tabular}
\end{sc}
\end{small}
\end{center}
\vskip -0.1in
\end{table}
        
\begin{figure}[ht]
        \centering
        \includegraphics[width=0.5\textwidth]{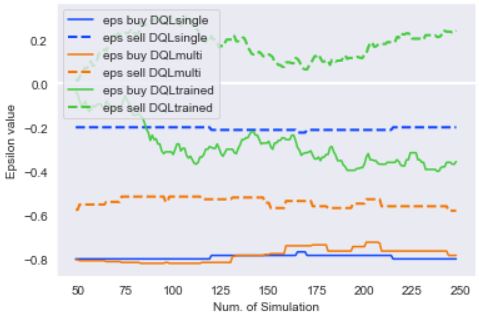}
        \caption{Epsilon evolution in single experiment - (50 sims. rolling window).}
        \label{fig:single4}
        \end{figure}

Regarding to strategies and epsilons, and analysing one of five rounds in current experiment, we notice how epsilons of three competing agents evolve along simulations. In  \autoref{fig:single4} it is noticeable how three different strategies comes up, depending on the agent. DQL-MM single-agent and DQL-MM multi-agent have similar buy epsilons, and it is in sell epsilons where the differences appear. At this point, DQL-MM single-agent (blue) has a less negative epsilon, what indicates that it prefers to earn less sell trades, but more profitable ones. However, DQL-MM multi-agent (orange) tries to have more sell trades, earning less reward per operation. Both pre-trained policies look very stable. On the other side, training DQL-MM agent (green) evolves its strategy from starting point, to adapt to the challenging environment. It learns a completely different policy, having even positive sell epsilons. This is far from pre-trained agents strategies. 

        We can conclude from this experiments that pre-trained policies are more volatile in terms of rewards than training ones. Furthermore, pre-trained policies are  impacted by the addition of new RL agents. All of this means that we should not look for profitable prefixed policies, but for profitable adaptive strategies.

\section{Conclusions}
\label{conclusions}

 Market makers play a key role in Financial Markets, where multiple participants operate under different strategies and goals. The use of market simulators such as ABIDES allow us to launch multiple experiments with different market structures and parameters. RL, meanwhile, has been proven as a promising technique to achieve market maker optimization problems. Regarding to the research, both experiments (trading as single DQL-MM agent, and competing among others DQL-MM) show good performance in terms of rewards and consistency. RL market maker agents behave quite good comparing to random agents. It is clear how, after some learning simulations, RL agents start to improve their profits, adapting their strategy in order to achieve as much reward as possible with available information. Furthermore, the strategy adaptation is even more evident when other RL agents are included. It is also interesting to remark how direct transfer learning can help us to migrate one profitable strategy to other competitive environment, while achieving good results. Related to this, it is worth mentioning that every profitable pre-trained agent will have to deal with new incoming agents, as environment becomes less stationary, no matter how good is the pre-trained policy it could have learned before. From a risk management point of view, pre-trained policies have shown more volatility in terms of average returns than learning ones, what indicates it is better to have agents that continuously learn from the environment than prefixed strategies. From a financial point of view, this experiments open us a door to deep dive in profitable market maker strategies in competitive scenarios based on RL, not only from market maker perspective but also from other market approaches.

\section*{Acknowledgements}

This research was funded in part by JPMorgan Chase Co. Any views or opinions expressed herein are solely those of the authors listed, and may differ from the views and opinions expressed by JPMorgan Chase Co. or its affiliates. This material is not a product of the Research Department of J.P. Morgan Securities LLC. This material should not be construed as an individual recommendation for any particular client and is not intended as a recommendation of particular securities, financial instruments or strategies for a particular client. This material does not constitute a solicitation or offer in any jurisdiction. This work has also been supported by the Madrid Government (Comunidad de Madrid-Spain) under the Multiannual Agreement with UC3M in the line of Excellence of University Professors (EPUC3M17), *and in the context of the V PRICIT (Regional Programme of Research and Technological Innovation). Finally, Javier García is partially  supported by the Comunidad de Madrid funds under the project 2016-T2/TIC-1712.
\bibliographystyle{ACM-Reference-Format}
\bibliography{export}

\appendix

\end{document}